%% file: latex/acl_latex.tex
\pdfoutput=1

\documentclass[11pt]{article}

\usepackage[preprint]{acl}

\usepackage{times}
\usepackage{latexsym}

\usepackage[T1]{fontenc}

\usepackage[utf8]{inputenc}

\usepackage{microtype}

\usepackage{inconsolata}

\usepackage{graphicx}
\usepackage{caption}
\usepackage{subcaption}
\usepackage{booktabs}
\usepackage{amsmath}
\usepackage{pifont}
\usepackage{multirow}
\usepackage{array,multirow}
\usepackage{array}
\newcolumntype{C}[1]{>{\centering\arraybackslash}p{#1}}

\title{SeaLLMs-Audio: Large Audio-Language Models for Southeast Asia}

\author{
  \quad Chaoqun Liu\; 
  \quad Mahani Aljunied\;
  \quad Guizhen Chen\;
  \quad Hou Pong Chan\; \\
  \quad \textbf{Weiwen Xu}\; 
  \quad \textbf{Yu Rong}\;
  \quad \textbf{Wenxuan Zhang}\thanks{\ Wenxuan Zhang is the corresponding author.} \\
  DAMO Academy, Alibaba Group\\ 
  \texttt{wxzhang@sutd.edu.sg}\\ \\
  \url{https://damo-nlp-sg.github.io/SeaLLMs-Audio/} 
}

\begin{document}
\maketitle
\begin{abstract}
We introduce \textbf{SeaLLMs-Audio}, the first large audio-language model (LALM) tailored for multiple Southeast Asian (SEA) languages—Indonesian (id), Thai (th), and Vietnamese (vi)-alongside English (en) and Chinese (zh). Trained on a large-scale audio corpus, SeaLLMs-Audio exhibits strong performance across diverse audio-centric tasks, spanning fine-grained audio understanding and voice-based interaction. Its key features include: 1) Multilingual: the model primarily supports 5 languages, namely Indonesian, Thai, Vietnamese, English, and Chinese; 2) Multimodal: the model accepts flexible input modalities, including audio only, text only, as well as audio with text; 3) Multi-task: the model supports a wide range of tasks, including audio analysis tasks such as Audio Captioning, Automatic Speech Recognition, Speech-to-Text Translation, Speech Emotion Recognition, Speech Question Answering, and Speech Summarization. It also enables voice-based dialogue, including answering factual, mathematical, and general knowledge queries. As a significant step towards advancing audio LLMs in Southeast Asia, we expect SeaLLMs-Audio to benefit both the regional research community and industry. To automate LALM evaluation for Southeast Asia, we introduce \textbf{SeaBench-Audio}, a benchmark spanning multiple tasks. Experiments show that SeaLLMs-Audio achieves competitive performance compared with other LALMs on SEA languages.~\footnote{SeaLLMs-Audio is publicly available at \url{https://github.com/DAMO-NLP-SG/SeaLLMs-Audio}}
\end{abstract}

\input{iclr2026/sections/introduction}

\input{iclr2026/sections/seallms_audio}
\input{iclr2026/sections/seabench_audio}

\input{iclr2026/sections/experiment}

\section{Conclusion}
In this study, we introduce a large audio-language model specifically designed for Southeast Asian languages, named SeaLLMs-Audio. Trained on an extensive multilingual audio corpus, SeaLLMs-Audio exhibits robust audio understanding and generation capabilities across Indonesian, Thai, and Vietnamese. To systematically assess LALMs within this region, we construct the SeaBench-Audio benchmark, encompassing multiple-choice and open-ended questions spanning 14 distinct tasks. Experimental outcomes highlight the strong performance of SeaLLMs-Audio on the proposed benchmark. We anticipate that SeaLLMs-Audio and SeaBench-Audio will promote further research on LALMs for Southeast Asia and stimulate broader efforts toward supporting low-resource languages.

\section{Limitation}
Due to limitations in manpower and computational resources, we confined SeaLLMs-Audio and SeaBench-Audio to three selected Southeast Asian languages. Nevertheless, the proposed methodology can be easily extended to a broader range of languages. Although SeaLLMs-Audio demonstrated strong performance, instances of language mixing still exist, a behavior commonly observed in other LALMs. We anticipate that this issue can be mitigated through reinforcement learning, which we identify as a promising direction for future work.


\subsubsection*{Acknowledgments}
We would like to express our special thanks to our professional, native linguists, Tantong Champaiboon, Nguyen Ngoc Yen Nhi and Tara Devina Putri, who helped build, evaluate, and fact-check our SeaBench-Audio dataset as well as evaluating our models across different aspects. We sincerely appreciate the valuable suggestions from Hao Zhang (DAMO Academy, Alibaba Group) on improving SeaLLMs-Audio.

\bibliography{iclr2026/iclr2026_conference,custom}

\appendix
\onecolumn
\input{iclr2026/sections/appendix}



\end{document}

%% file: iclr2026/sections/introduction.tex
\section{Introduction}
Large audio-language models (LALMs) \citep{chu_qwen-audio_2023,chu_qwen2-audio_2024,he_meralion-audiollm_2024,pipatanakul_typhoon_2024,held_distilling_2024} have shown impressive capabilities in understanding the rich information contained in audio signals. However, most existing LALMs support only one or two languages, most typically English, leaving multilingual and low-resource regions under-represented.

In Southeast Asia (SEA), significant progress has been made in developing multilingual large language models (LLMs), such as SeaLLMs \citep{nguyen_seallms_2024,zhang_seallms_2025,zhao_babel_2025}, Sailor \citep{dou_sailor_2024,dou_sailor2_2025}, and SEA-LION\footnote{\url{https://huggingface.co/aisingapore/collections}} series. Despite their multilingual reach, these models operate solely in the textual modality and lack the ability to process audio inputs—an essential component of natural human communication.

\begin{figure*}[t!]
    \centering
    \includegraphics[width=0.9\linewidth]{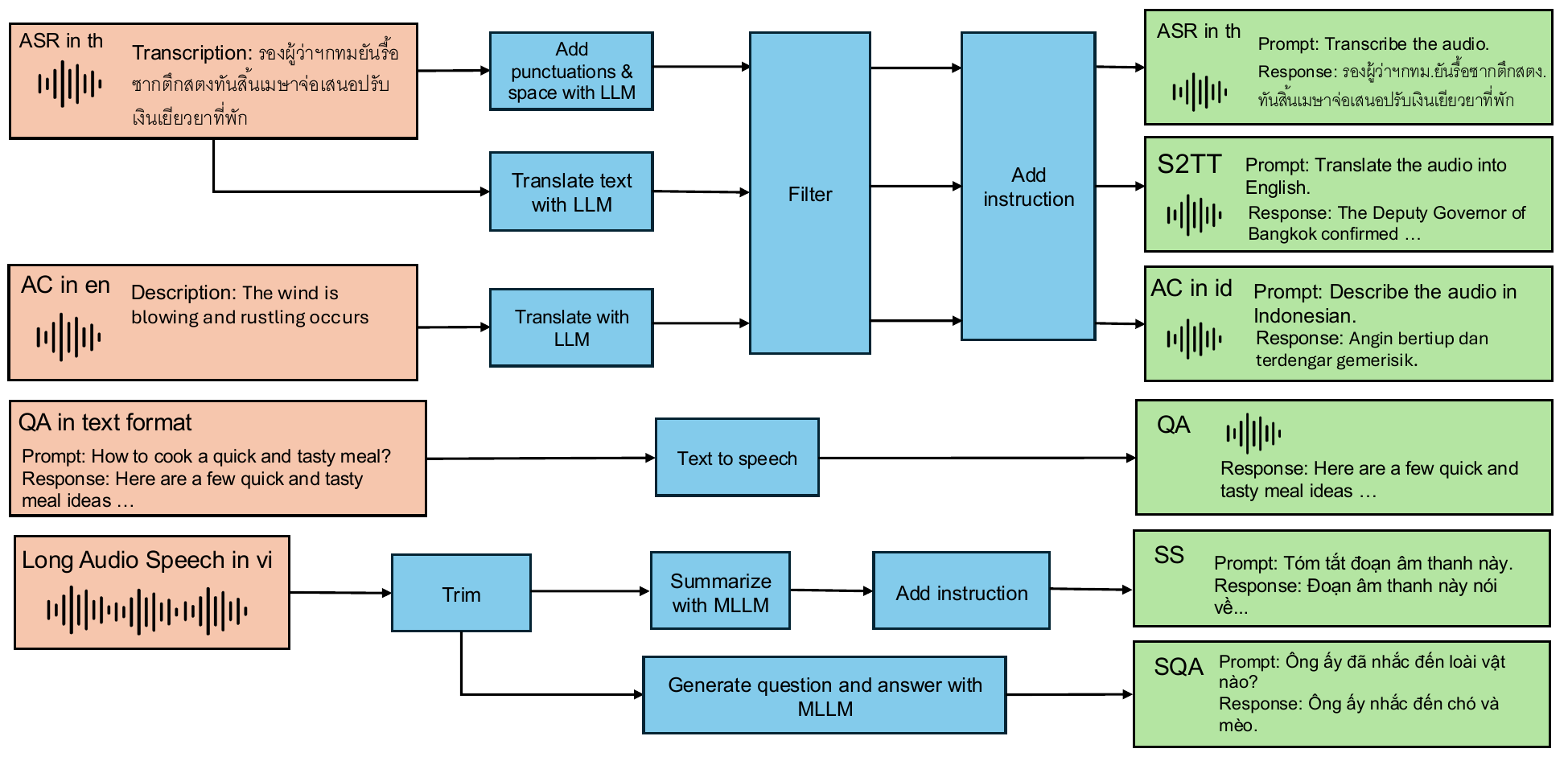}
    \caption{Illustration of data curation process for SeaLLMs-Audio. 
    }
    \label{fig:training_data_curation}
\end{figure*}

Beyond the absence of LALMs tailored for Southeast Asian languages, progress is further constrained by the lack of comprehensive and rigorous evaluation frameworks. Existing benchmarks, such as SeaEval \citep{wang_seaeval_2024}, SeaExam, and SeaBench \citep{liu_seaexam_2025}, focus primarily on textual evaluation within SEA contexts. Meanwhile, audio-related benchmarks remain limited to specific tasks like automatic speech recognition (ASR) \citep{wang_audiobench_2025}, without providing a holistic assessment of audio understanding and voice-based interaction. This lack of broad, multimodal benchmarks continue to impede the advancement of audio-language modeling in SEA languages.

To bridge above gaps, we introduce SeaLLMs-Audio (Southeast Asian Large Language Models with audio capabilities), a Large Audio-Language Model designed specifically for Southeast Asia. SeaLLMs-Audio is trained using data from a comprehensive curation pipeline that aggregates, organizes, and synthesizes multimodal resources across SEA languages, as illustrated in Figure~\ref{fig:training_data_curation}. The curated dataset spans diverse tasks, including automatic speech recognition (ASR), audio captioning (AC), speech-to-text translation (S2TT), speech summarization (SS), audio question answering (AQA), and multimodal reasoning.

Furthermore, to facilitate standardized evaluation, we present SeaBench-Audio, a manually curated benchmark for assessing LALMs in Southeast Asian languages. SeaBench-Audio encompasses multiple open-ended task categories that reflect real-world, multimodal language understanding scenarios. To facilitate consistent and scalable evaluation, we adopt an LLM-as-a-judge framework with task-specific prompt templates, achieving high agreement with human annotations. Experimental results on SeaBench-Audio demonstrate that SeaLLMs-Audio delivers robust and competitive performance across a wide range of audio-language tasks.

Our key contributions are as follows.
\begin{itemize}
    \item We present \textbf{SeaLLMs-Audio}, a large-scale audio--language model specifically designed for Southeast Asian contexts.
    \item We develop \textbf{SeaBench-Audio}, a comprehensive benchmark dedicated to evaluating LALMs within the SEA region.
    \item Our experimental analyses indicate that \textbf{SeaLLMs-Audio} achieves strong performance on the \textbf{SeaBench-Audio} benchmark.
\end{itemize}

%% file: iclr2026/sections/seallms_audio.tex
\section{SeaLLMs-Audio}
In this section, we illustrate the training data curation pipeline followed by the model architecture. 

\subsection{Comprehensive Data Curation Pipeline}
This training dataset for SeaLLMs-Audio contains 1.58M conversations for multiple tasks, including 7\% multi-turn dialogues that better reflect real-world interactive scenarios. The tasks can be roughly classified as the following categories: automatic speech recognition (ASR), audio captioning (AC), speech-to-text translation (S2TT), question answering (QA), speech summarization (SS), audio question answering (AQA), chat, math, and factoid QA (fact) and other tasks (mixed). 

The training dataset was curated from multiple data sources, including public datasets and private data. Public datasets include: GigaSpeech \citep{chen_gigaspeech_2021}, GigaSpeech2 \citep{yang_gigaspeech_2025}, Common Voice \citep{ardila_common_2020}, AudioCaps \citep{kim_audiocaps_2019}, VoiceAssistant-400 \citep{xie_mini-omni_2024}, YODAS2 \citep{li_yodas_2024}, and Multitask National Speech Corpus \citep{he_meralion-audiollm_2024}. 
As these datasets span multiple sources with disparate formats (e.g., different audio encodings, annotation schemas, and text structures), they cannot be directly used for end-to-end training. We therefore perform comprehensive preprocessing to unify the data.
Figure \ref{fig:training_data_curation} shows the overall data curation pipeline with some examples.
The following describes the construction process for each task.
\paragraph{ASR} For ASR datasets such as GigaSpeech, we normalize transcripts to improve readability. For example, we transform \texttt{"AND LOOK AT THE PERCENTAGE OF REPORTS <PERIOD>"} into \texttt{"And look at the percentage of reports."} For GigaSpeech2, which includes Thai, Indonesian, and Vietnamese while its text does not contain punctuations, we employ a selected LLM to restore punctuations and spacing, producing more reader-friendly text for each language. As LLMs may introduce errors, we discard samples whose outputs are inconsistent with the original transcripts. 
\paragraph{S2TT} Given the absence of open-source S2TT datasets for SEA languages, we construct such data by leveraging ASR corpora in different languages.
More specifically, since each unit of the ASR data comprises the same-language speech audio plus its text transcription, we utilize this text to create translations into multiple targeted languages. This results in data pairs of speech audio in one language plus their translated text in another language.
\paragraph{AC} AudioCaps provides captions exclusively in English. To accommodate Southeast Asian languages, we translate these captions into the respective target languages.
\paragraph{QA} This set is curated to obtain audio questions with text answers. To do this, we make use of existing question-answer pairs in text format. The answers are kept unchanged in text form, while the text questions are converted into audio with text-to-speech (TTS) models. No translation is involved. After manually assessing samples of the quality of TTS outputs from several models, we finally select Google Text-to-Speech \footnote{\url{https://cloud.google.com/text-to-speech}}.
\paragraph{SS} In order to curate SS dataset, we sample a piece of speech audio from YODAS2 dataset and ask Gemini-2.0-Flash to summarize it in a specified language.
\paragraph{AQA} In order to create natural questions and audio about a piece of audio, we first sample a piece of audio from the YODAS2 dataset, which contains audio for YouTube videos. After that, we prompt Gemini-2.0-Flash to generate a question about the audio and provide the corresponding answer. 
\paragraph{chat} In order to create voice chat data, we make use of existing text conversation data and convert the user input into audio format with Google TTS. As Google TTS has only a few voice types for each language, we also transcribe part of the data with \texttt{gpt-4o-mini-tts} to improve the diversity.
\paragraph{math and fact}
For math and factoid QA instruction data, we also transcribe the prompts into speech with Google TTS.

\begin{figure}[ht!]
    \centering
    \begin{subfigure}[b]{0.45\textwidth}
        \centering
        \includegraphics[width=\textwidth]{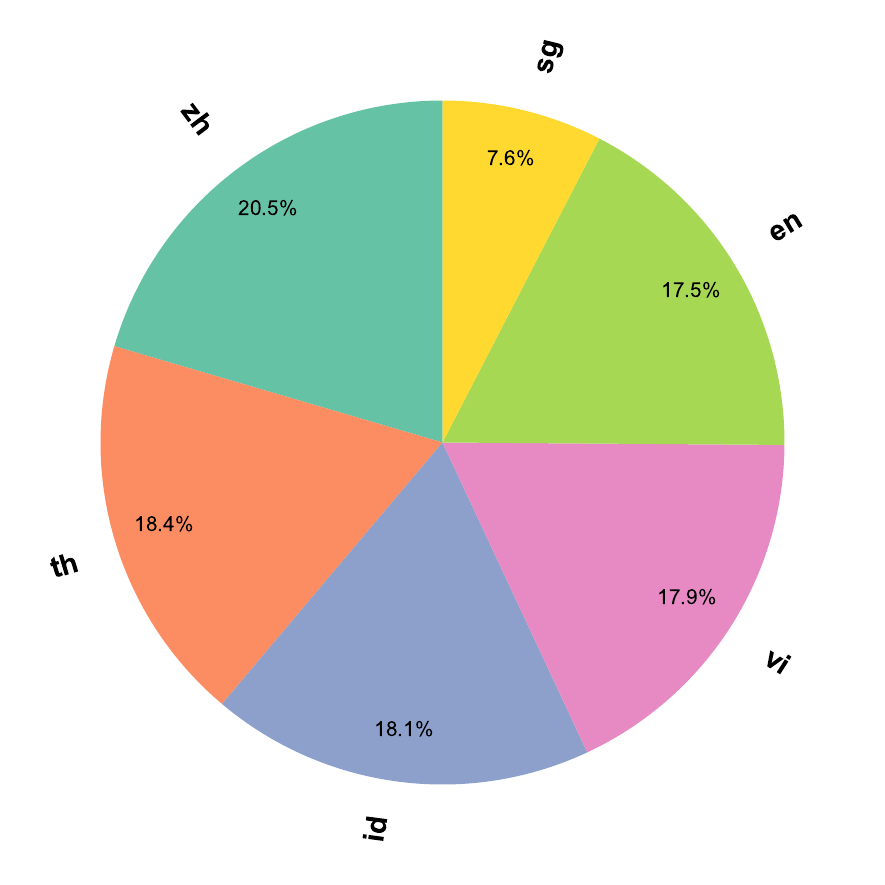}
        \caption{Data Distribution by Task Type}
    \end{subfigure}
    \begin{subfigure}[b]{0.45\textwidth}
        \centering
        \includegraphics[width=\textwidth]{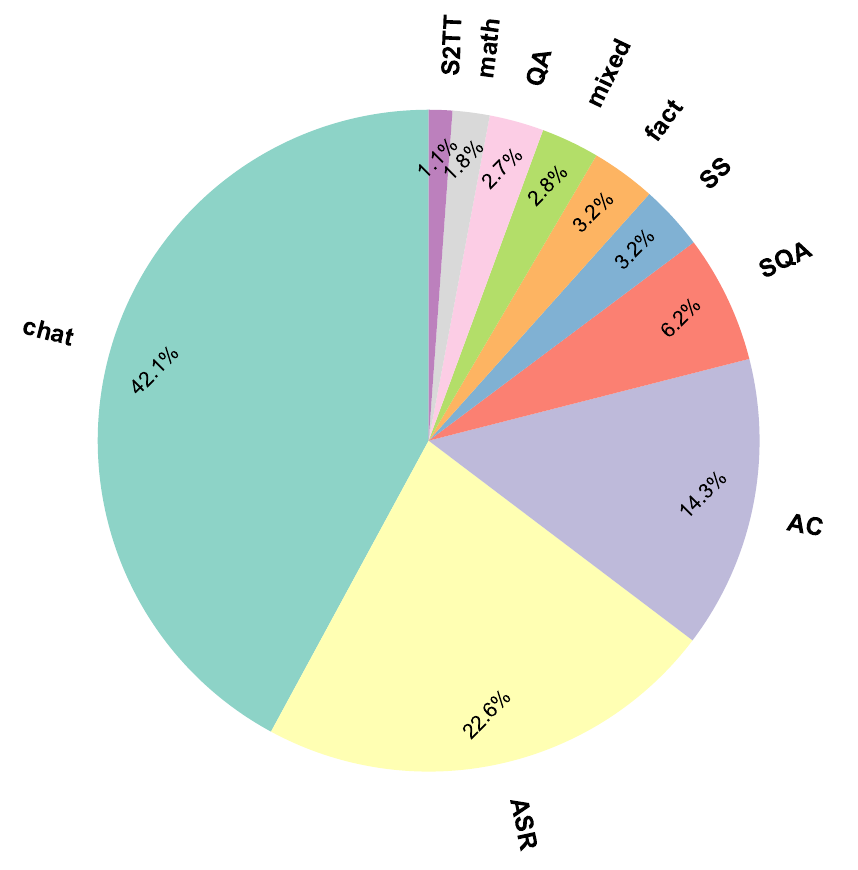}
        \caption{Data Distribution by Language}
    \end{subfigure}
    
    \caption{Training data distribution across (a) languages and (b) task types.}
    \label{fig:training_data}
\end{figure}

The distribution of the training data by language and by task type is shown in Figure \ref{fig:training_data}.

\subsection{Model Architecture}
\begin{figure}[t]
  \centering
  \includegraphics[width=0.95\linewidth]{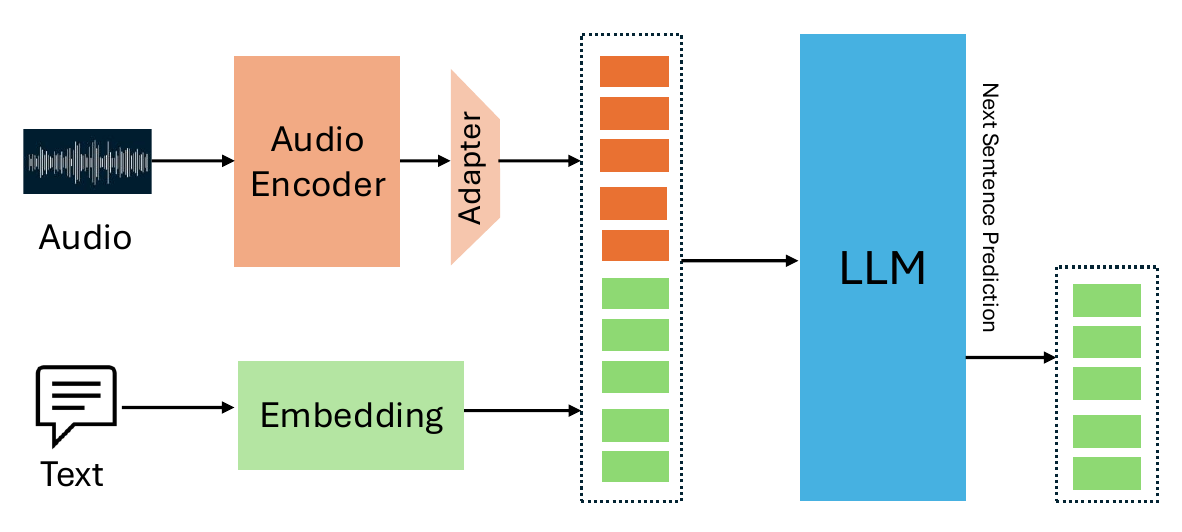}
  \caption{Architecture of SeaLLMs-Audio.}
  \label{fig:architecture}
\end{figure}

SeaLLMs-Audio builts upon Qwen2-Audio-7B \citep{chu_qwen-audio_2023} and Qwen2.5-7B-Instruct \citep{qwen_qwen25_2025}. The architecture is shown in Figure \ref{fig:architecture}. We replace the LLM module in Qwen2-Audio-7B \citep{chu_qwen2-audio_2024} by Qwen2.5-7B-Instruct \citep{qwen_qwen25_2025}. In this way, we harness the advantages of both models: Qwen2-Audio-7B audio encoder can encode the audio features for speech and non-speech audios effectively, and Qwen2.5-7B-Instruct has strong multilingual capabilities. Due to the hidden embedding mismatch, the audio adapter is newly initialized. After that, we do full-parameter fine-tuning on our newly curated large-scale audio dataset, which contains multiple tasks. Given paired data $(a, x)$, with $a$ denoting the audio sequences and $x$ denoting the optional corresponding text sequences, the training objective is to maximize the likelihood of the subsequent text token, formulated as
\begin{equation}
\mathcal{P}_{\theta}(x_t \mid x_{<t}, a),
\label{eq:conditional_prob}
\end{equation}
conditioned on the audio representations and the preceding text tokens $x_{<t}$, where $\theta$ represents the trainable parameters of the LALM. We train the model on the dataset for 1 epoch, which took ~6 days to complete on 32 A800 GPUs.

%% file: iclr2026/sections/seabench_audio.tex
\section{SeaBench-Audio}
Due to the absence of standard audio benchmarks for evaluating audio LLMs in SEA languages, we manually create a benchmark called SeaBench-Audio. It comprises 14 tasks: 1) Tasks with both audio and text inputs:
 Automatic Speech Recognition (ASR), Speech-to-Text Translation (S2TT), Speech Summarization (SS), Speech Question Answering (SQA), Customer Service (CS), Safety, Audio Cationing (AC), Audio Question Answering (AQA), Speaker Identifiers (SKI), and Speech Emotion Recognition (SER); 2) Tasks with only audio inputs: Life, Medical (MED), Math, and Fact. 
The task descriptions and annotation criteria are summarized in Table \ref{tab:seabench_details} in Appendix \ref{sec:appendix}. 

\begin{figure}[t!]
  \centering
  \includegraphics[width=0.98\linewidth]{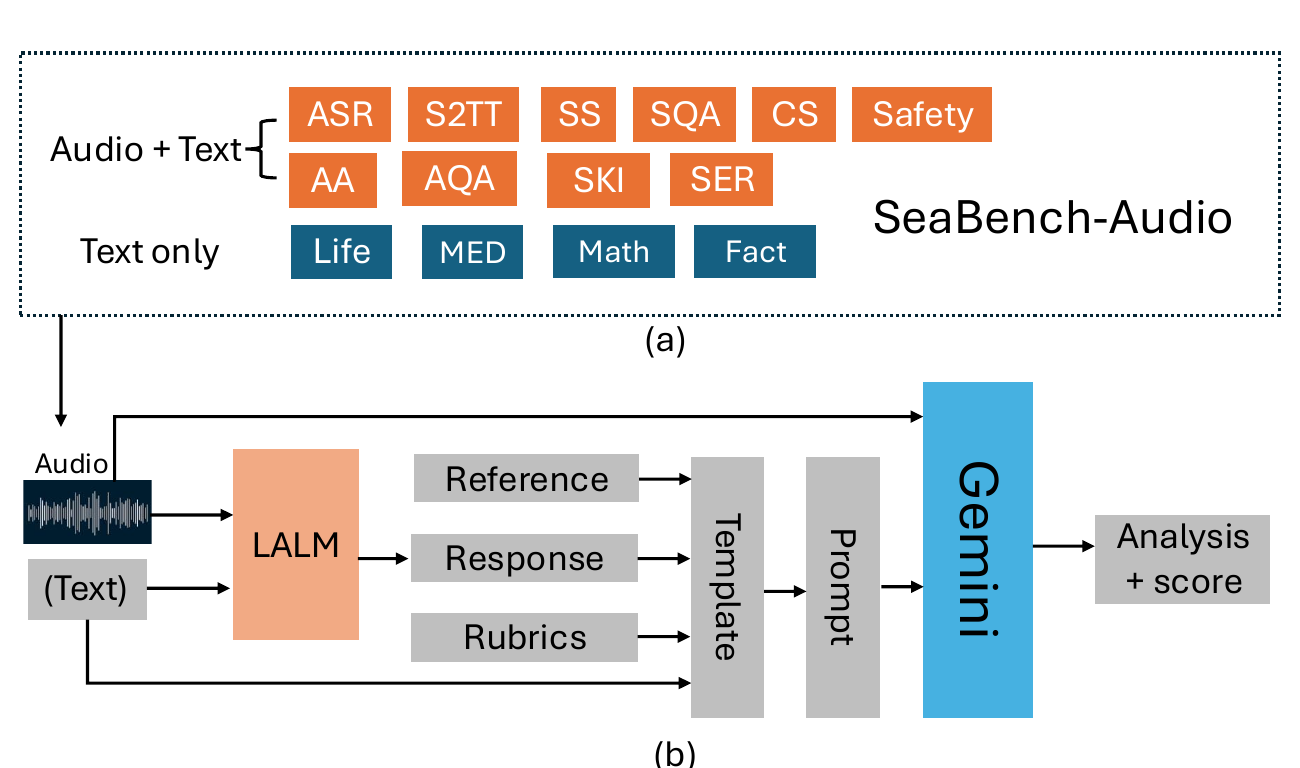}
  \caption{(a) An overview of the task in SeaBench-Audio (b) Evaluation pipeline with LLM-as-a-judge framework.}
  \label{fig:seabench_audio}
\end{figure}

An overview of the datasets is provided in Figure \ref{fig:seabench_audio}(a). For each language, we engage a professional native linguist to annotate 10 questions per task. 
One exception is S2TT task, which requires the translation between two languages. 
For id/th/vi, we construct two versions—native audio to English text and English audio to native text—each with 10 questions. For English, we omit this task to prevent redundancy. Consequently, there are 150 questions for each SEA language and 130 for English, yielding a total of 580 questions. For every question, a linguist supplies a reference to facilitate scoring. The benchmark underwent multiple rounds of careful review to ensure quality.

For evaluation, qualified native speakers rated each response on a scale of 1 to 5, with 5 representing the highest quality. However, human evaluations are expensive and time-consuming. In order to facilitate automatic evaluation, we employ an LLM-as-a-judge framework. We choose Gemini-2.5-flash (Gemini) \citep{comanici_gemini_2025}, due to its capabilities of audio understanding and good balance of cost and performance. As shown in \ref{fig:seabench_audio}(b), the procedures to evaluate an LALM are: 1) Generate responses for each instance with the LALM; 2) Construct an evaluation prompt with the text instruction (optional), reference answer, response, rubrics, and the template; 3) Prompt Gemini with the audio and evaluation prompt; 4) Extract the score from the final response. The prompt template for LLM-as-a-judge is shown in Figure \ref{fig:llm-judge-prompt} in Appendix \ref{sec:appendix}. For each task, we additionally engage linguists to develop a task-specific evaluation rubric for the responses on a scale of 1 to 5.
We hypothesize that tasks exhibit distinct characteristics, and a dedicated rubric more accurately captures the nuances of each task.

%% file: iclr2026/sections/experiment.tex
\section{Experiments}
We compare the performance of SeaLLMs-Audio with relevant LALMs with similar sizes, including: 
\paragraph{Qwen2-Audio-7B-Instruct (Qwen2-Audio)} This is the latest version of Qwen-Audio series \citep{chu_qwen2-audio_2024}, which mainly focus on English and Chinese. It shares the same base audio encoder as SeaLLMs-Audio.
\paragraph{Qwen2.5-Omni-7B (Qwen2.5-Omni)} This is a multimodal model that perceive diverse modalities including text, audio, image, and video \citep{xu_qwen25-omni_2025}. It also adopted the audio encoder from Qwen2-Audio. In this work, we only compare its performance with audio input.
\paragraph{MERaLiON-AudioLLM-Whisper-SEA-LION (MERaLiON)} This model was trained to understand Singlish, which was trained on 62 million multimodal instruction samples \citep{he_meralion-audiollm_2025}. Since it was trained with both audio and text instructions, we add a text instruction \textit{"Please follow the instruction in the speech."} for tasks with no text input.

\paragraph{MERaLiON-2-10B (MERaLiON-2)} This is a concurrent work with SeaLLMs-Audio. Compared with MERaLion, it supports more languages, including English, Chinese, Indonesian, Thai, and Vietnamese; thus, it has a similar motivation to SeaLLMs-Audio \citep{he_meralion-audiollm_2025}. Like MERaLiON, we add a text instruction \textit{"Please follow the instruction in the speech."} for tasks without text input.

All the LALMs can accept audio with text as input. In order to evaluate the quality of LLM-as-a-judge, we also conducted human evaluations. The judging criteria are shown in Table \ref{tab:human_rules} in Appendix \ref{sec:appendix}. We engage the benchmark annotators to do the evaluations as they are familiar with the benchmarks.

\subsection{Main Results}
\begin{figure}[t!]
    \centering
    \begin{subfigure}[b]{0.45\textwidth}
        \centering
        \includegraphics[width=\textwidth]{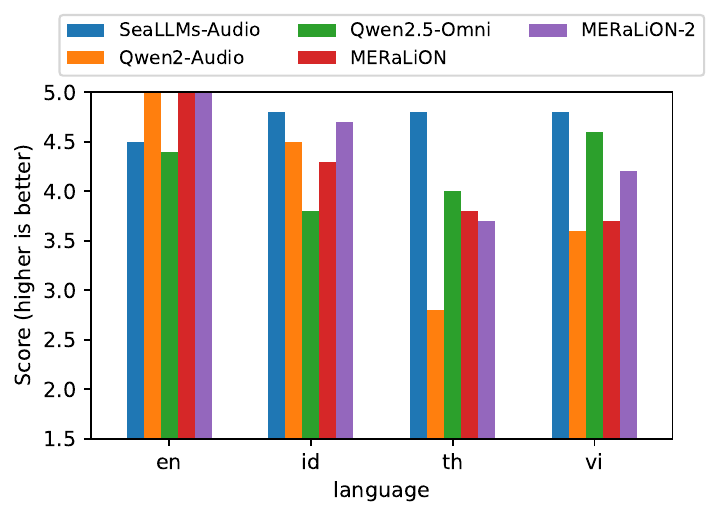}
        \caption{}
    \end{subfigure}

    \begin{subfigure}[b]{0.45\textwidth}
        \centering
        \includegraphics[width=\textwidth]{iclr2026/figs/results/scores_lang_human_languageQuality.pdf}
        \caption{}
    \end{subfigure}
    
    \caption{Performance of the models on SeaBench-Audio accessed by human evaluators: (a) average scores for overall performance, and (b) average scores for output language quality. Each response is evaluated on a 1–5 scale, with 5 indicating the highest quality. Human evaluations were performed blind, without disclosure of the generating model.}
    \label{fig:main_results} 
\end{figure}

\begin{figure}[htbp]
  \centering
  \includegraphics[width=0.98\linewidth]{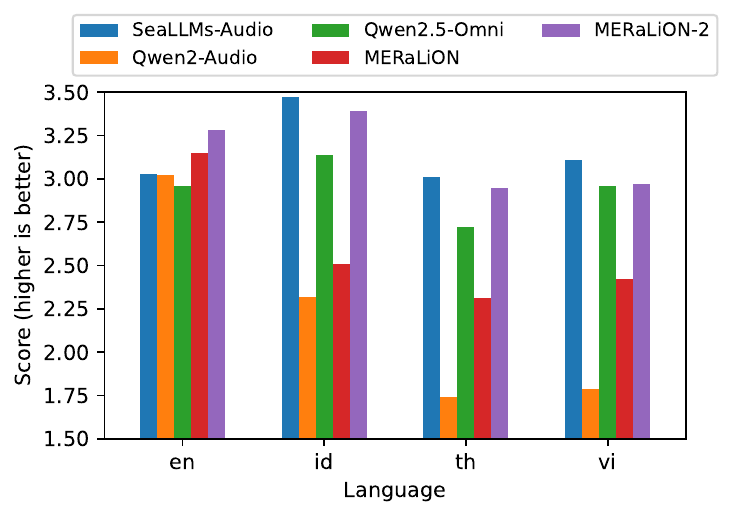}
  \caption{Average scores of the models on SeaBench-Audio accessed by Gemini-2.5-flash.}
  \label{fig:scores_langu_llm}
\end{figure}

Figure \ref{fig:main_results} shows the human evaluations. Evaluators assessed both overall performance and language quality, the latter referring to the correctness of language usage in responses. Language quality was rated on a 1-5 scale, where 5 indicated entirely correct language devoid of code-switching. We can see that SeaLLMs-Audio achieves the best language quality for the three SEA languages. 
The LLM-as-a-judge evaluation result is shown in Figure \ref{fig:scores_langu_llm}. 
From these results, SeaLLMs-Audio attains the strongest performance on id/th/vi, irrespective of whether evaluation is conducted by human annotators or Gemini. Additional observations include: (1) MERaLiON-2 surpasses MERaLiON, which is expected given that MERaLiON-2 is the newer iteration; and (2) Qwen-Omni outperforms Qwen2-Audio across the three languages, consistent with its more recent release. This alignment with prior expectations supports the validity of the LLM-as-a-judge framework.

\subsection{Analysis}
To further understand the capabilities of SeaLLMs-Audio and SeaBench-Audio, we conduct more analysis on the results.

\paragraph{How does SeaLLMs-Audio perform on each task?}
\input{iclr2026/tables/scores_by_task}

In addition to assessing average performance across languages, we further examine model outcomes by task. Table \ref{tab:scores_by_task} presents average scores for each model across all evaluated tasks. Since the English subset excludes the S2TT task, averages for that setting are computed only over id, th, and vi. MERaLiON-2 consistently achieves the strongest results in audio comprehension tasks—specifically ASR, S2TT, and SER—which we attribute to its substantially larger and more diverse training corpus \citep{he_meralion-audiollm_2025}. Conversely, SeaLLMs-Audio attains state-of-the-art performance in selected categories, including \textit{fact, life, MED}, and \textit{math}. We ascribe SeaLLMs-Audio’s advantages to the extensive scope and heterogeneity of its training data, encompassing both varied task types and multimodal input formats.

\begin{figure}
    \centering
    \includegraphics[width=0.9\linewidth]{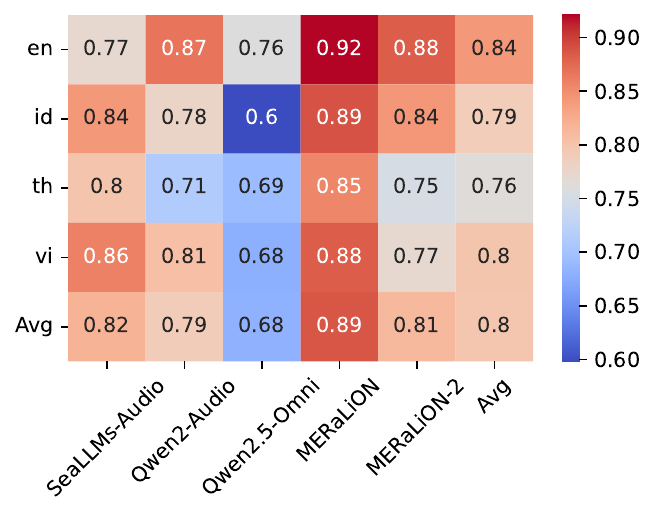}
    \caption{The Pearson correlation coefficient between human judgements and LLM judgements.}
    \label{fig:correlation}
\end{figure}

\paragraph{How is LLM-as-a-judge consistent with human judges?}
We observed that the scores by human judgments and LLM-as-a-judge evaluations are not perfectly aligned. To evaluate their correlation, we calculate their Pearson correlation coefficient. As shown in Figure \ref{fig:correlation}, LLM-as-a-judge and human judges have  an average correlation coefficient of 0.8, which shows high correlation between the scores by humans and by the LLM judge.
We also calculate the agreement between human judges and LLM judges when comparing the responses from two models. As shown in Table \ref{tab:agreement}, they have an average agreement of 69\% with tie and 93\% without tie, which is even higher than the result in MT-bench \cite{zheng_judging_2023}. Such high agreement between humans and the LLM judge shows the reliability of SeaBench-Audio.

\begin{table}[ht!]
\centering
\small
\begin{tabular}{lccccc}
    \toprule
    \textbf{Setup} & \textbf{en} & \textbf{id} & \textbf{th} & \textbf{vi} &\textbf{Avg}\\
    \midrule
    w/ tie (R=33\%) & 68\% & 71\% & 68\% & 70\% & 69\% \\
    w/o tie (R=50\%) & 92\% & 95\% & 92\% & 94\% & 93\% \\
    \bottomrule
\end{tabular}
\caption{Agreement between human judges and the LLM judge. We convert the single-answer grading to pairwise comparison results for calculating the agreement. "w/ tie" includes tie scores and non-tie scores. "w/o tie" includes only non-tie scores. "R=" indicates the agreement between two random judges.}
\label{tab:agreement}
\end{table}

%% file: iclr2026/tables/scores_by_task.tex
\begin{table*}[ht!]
    \centering
    \setlength\tabcolsep{2pt}
    \small
    \begin{tabular}{lp{0.08\textwidth}p{0.08\textwidth}p{0.08\textwidth}p{0.08\textwidth}p{0.08\textwidth}cp{0.08\textwidth}p{0.08\textwidth}p{0.08\textwidth}p{0.08\textwidth}p{0.08\textwidth}}
    \toprule
    \multirow{2}{*}{Task} 
    & \multicolumn{5}{c}{\textbf{Human-as-a-judge}} 
      & \multicolumn{1}{c}{} 
      & \multicolumn{5}{c}{\textbf{LLM-as-a-judge}} \\
      \cmidrule(lr){2-6} \cmidrule(lr){8-12}
        task & SeaLLMs-Audio & Qwen2-Audio & Qwen2.5-Omni & MERa-LiON & MERa-LiON-2 & ~ & SeaLLMs-Audio & Qwen2-Audio & Qwen2.5-Omni & MERa-LiON & MERa-LiON-2 \\ 
        \midrule
        AA & \textbf{2.4} & 2.1 & \textbf{2.4} & 2.2 & \textbf{2.4} & ~ & 2.3 & 2 & 2.3 & 2 & \textbf{2.4} \\ 
        AQA & 3.2 & 2.3 & 2.8 & 3.2 & \textbf{3.4} & ~ & 3.5 & 2.6 & 2.8 & 3.3 & \textbf{3.7} \\ 
        ASR & 3.9 & 2.2 & 3.3 & 2.9 & \textbf{4.2} & ~ & 3.7 & 1.7 & 2.8 & 2.8 & \textbf{4.4} \\ 
        CS & 3.3 & 2.7 & 2.9 & 3.2 & \textbf{3.6} & ~ & 3.2 & 2.5 & 3.3 & 3.3 & \textbf{4} \\ 
        MED & \textbf{3.6} & 1.8 & 3.3 & 1.2 & 1.6 & ~ & \textbf{3.5} & 1.5 & 2.8 & 1 & 1.6 \\ 
        S2TT\_EX & 2.9 & 2.1 & 2.5 & 3.6 & \textbf{3.7} & ~ & 3.5 & 2.4 & 2.9 & 3.8 & \textbf{3.9} \\ 
        S2TT\_XE & 2.6 & 1.2 & 3.2 & 2.3 & \textbf{3.6} & ~ & 2.3 & 1.2 & 2.6 & 2.1 & \textbf{3.3}\\ 
        SER & 2.8 & 2.4 & 2.7 & 2.3 & \textbf{3.2} & ~ & 3.1 & 2 & 3.2 & 2.2 & \textbf{3.5} \\ 
        SKI & 2.2 & 2.6 & 1.9 & \textbf{3.3} & 2.8 & ~ & 2.5 & 2.9 & 2.3 & \textbf{3.3} & 2.7 \\ 
        SQA & 4.1 & 2.2 & 4 & 3.7 & \textbf{4.3} & ~ & 4.2 & 2.5 & 4.2 & 3.4 & \textbf{4.3} \\ 
        SS & 3.2 & 2.3 & 3.3 & 3.2 & \textbf{4.1} & ~ & 3.4 & 1.7 & 3.3 & 3.4 & \textbf{4.4} \\ 
        fact & \textbf{3.1} & 1.6 & 2.9 & 1.9 & 2 & ~ & \textbf{3.1} & 1.8 & 3 & 1.3 & 1.6 \\ 
        life & \textbf{3.3} & 1.9 & 3.1 & 1.2 & 1.3 & ~ & \textbf{3.7} & 1.7 & 3.1 & 1 & 1.2 \\ 
        math & \textbf{3.6} & 1.3 & 2.6 & 1.7 & 2.3 & ~ & \textbf{4} & 1.2 & 3 & 1.5 & 3.5 \\ 
        safety & 2 & 2.2 & 2.1 & 1.6 & \textbf{2.5} & ~ & 2.1 & 1.6 & \textbf{2.7} & 1.6 & 2.4 \\ 
        \bottomrule
    \end{tabular}
    \caption{Average scores for each task across the three SEA languages. We show the scores judged by humans and by Gemini-2.5-flash. The highest scores for each task are highlighted in \textbf{bold}.}
    \label{tab:scores_by_task}
\end{table*}

%% file: iclr2026/sections/appendix.tex
\section{Appendix}
\label{sec:appendix}

\input{iclr2026/tables/seabench_tasks}

\begin{figure*}[t!]
    \centering
    \includegraphics[width=0.95\linewidth]{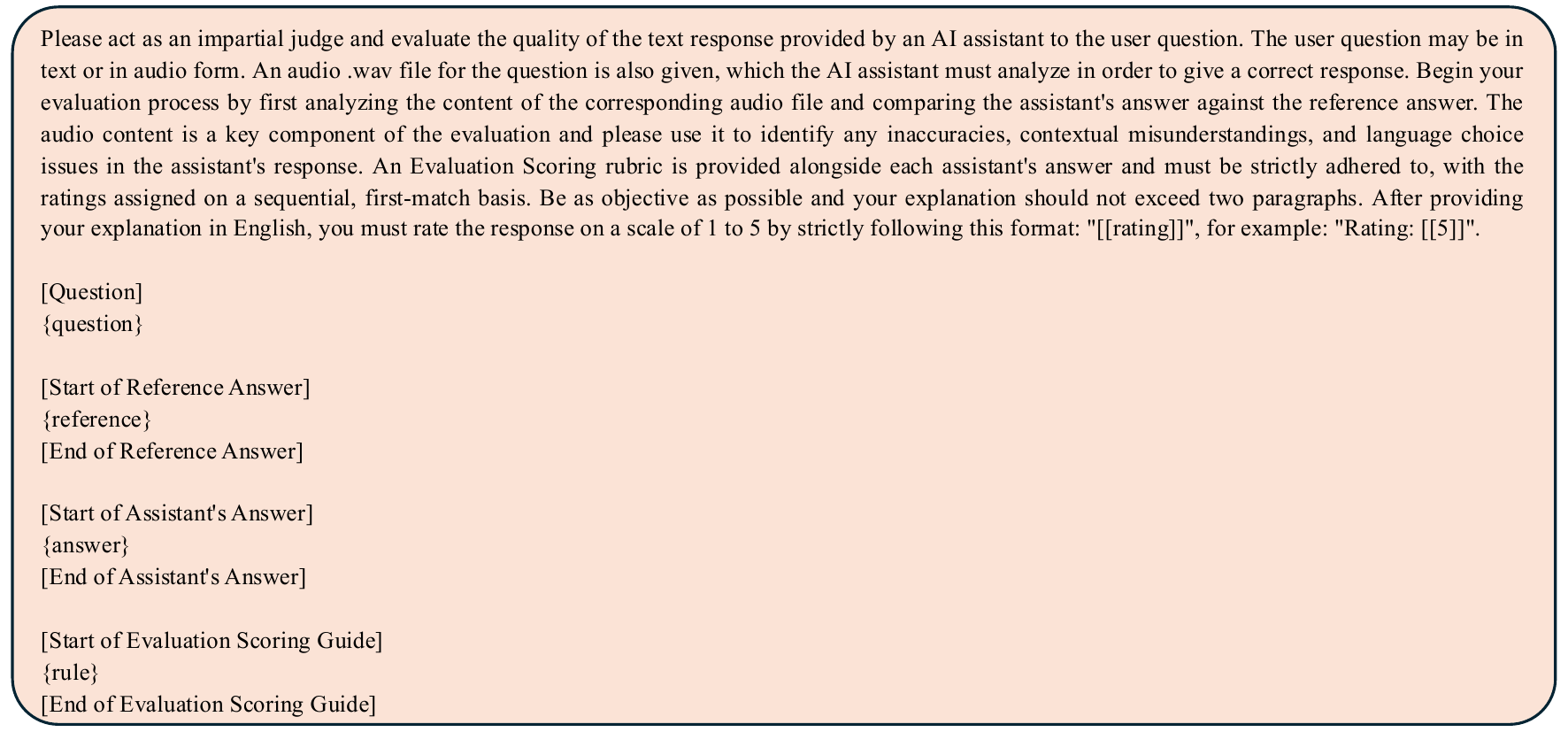}
    \caption{The prompt template for LLM-as-a-judge.}
    \label{fig:llm-judge-prompt}
\end{figure*}

\input{iclr2026/tables/human_rubrics}

%% file: iclr2026/tables/seabench_tasks.tex
\begin{table*}[ht!]
\centering
\footnotesize
{\fontsize{8}{9}\selectfont
\begin{tabular}{C{0.07\textwidth}p{0.75\textwidth}C{0.03\textwidth}}
    \toprule
    \textbf{Task} & \textbf{Task description and requirements} & \textbf{text?} \\
    \midrule
    ASR & Automatic Speech Recognition. For this task, each input consists of an audio file and a text instruction. Some clips feature regional accents, and every language in this category includes at least one example of informal speech. The set also contains multi-sentence audio clips. Each text instruction is uniquely phrased. & Yes \\
\midrule
SS & Speech Summarization. Each unit includes a text instruction to convert an audio file into a shorter text. The instruction may limit output length by word count, number of sentences, or by specifying a short format (e.g., headline or title). & Yes \\
\midrule
SER & Speaker Emotion and Sentiment Recognition. This category includes 5 emotion detection units and 5 sentiment detection units. Each unit’s text instruction provides options from which the model must choose the correct answer. Every audio clip contains speech whose lexical content does not reflect the speaker’s emotion or sentiment. This setup compels the model to rely on paralinguistic features to determine the correct answer.& Yes \\
\midrule
S2TT\_EX, S2TT\_XE & Speech to Text Translation. There are two speech translation categories in this task: English-to-local (EX) and local-to-English (XE). This category includes units from specialized domains (e.g., courtroom, military, medical, addressing royalty) across languages. Translation challenges—such as idioms, informal varieties, appropriate first- and second-person pronouns, and polysemy—are represented in these units. & Yes \\
\midrule
SQA & Speech Question Answering. This task evaluates a model’s ability to extract or infer specific information from an audio file. Each unit’s text instruction is crafted to elicit a precise answer for reliable evaluation. The targeted information may extend beyond named entities present in the audio. & Yes \\
\midrule
AC & Audio Captioning. Each audio unit includes non-speech sounds from both animate and inanimate sources, as well as natural or man-made scenes. The accompanying text prompts are crafted to elicit detailed descriptions of the entire audio content. & Yes \\
\midrule
AQA & Audio Question Answering. The audio input for AQA consists of non-speech snippets, including event sounds or sequences of related sounds from human activities or nature. The text instructions test a model’s abilities in object/event detection, sound tracking (e.g., duration), segmentation, and sequential analysis. Some prompts include multiple-choice options (e.g., “... takes place outdoors or indoors?”). & Yes \\
\midrule
SKI & Speaker Identifiers. This category focuses on non-linguistic aspects of audio (e.g., number of speakers, turn-taking, speaking rate) and speaker-related tasks such as gender, relative age, and accent or regional variety identification. & Yes \\
\midrule
Life & Life. The audio files in this set come from text questions posted on popular forums and social media in the respective SEA-language countries. Questions are carefully chosen to cover diverse topics, suit model prompting, and remain manageable for evaluation. Each question is recorded with a unique voice. There are no separate text instructions, as the audio explicitly contains the questions. & No \\
\midrule
CS & Customer Service. This set includes 10 customer care scenarios. Each audio clip features either a single customer or a dialogue between a customer and a CS officer. The scripts simulate real calls—for example, clarifying product or price information, checking delivery status, requesting refunds, or reporting faulty products. The instructions include 6 units with answer options and 4 open-ended units. These units are designed to assess fine-grained CS knowledge and were curated in consultation with a customer care professional. & Yes \\
\midrule
MED & Medical Patient Question. Each unit contains an audio recording of a patient describing their condition and requesting medical advice. These recordings are human-voiced readings of questions sourced mainly from publicly available hospital websites, with some from open-source Q\&As. This category evaluates a model’s ability to respond within a specific medical domain. References are provided by doctors from the respective hospitals or are human-verified. & No \\
\midrule
Safety & Safety. Containing an audio file and a text instruction, each unit in this category is designed to provoke an unsafe response from models. The set includes both country-specific safety violations and universally unsafe topics. Each unit is curated so that a safe, desired response requires analyzing the audio content—the text instruction alone does not reveal anything unsafe. This ensures that any model rejection stems from understanding the speech/audio elements. & Yes \\
\midrule
Math & Math. There are no written instructions in this set; each audio clip contains a complete math question. The questions span various math topics for grades 7–12. & No \\
\midrule
Fact & Fact. Audio clips in this set pose explicit factual questions across diverse topics. No text instructions accompany the audio files. Subjects include history, economics, medicine, technology, and more. & No \\
    \bottomrule
\end{tabular}
}
\caption{Verbalizers for the evaluation datasets.}
\label{tab:seabench_details}
\end{table*}

%% file: iclr2026/tables/human_rubrics.tex
\begin{table*}[ht!]
\centering
\small
\begin{tabular}{C{0.07\textwidth}p{0.75\textwidth}}
    \toprule
    \textbf{Score} & \textbf{Criteria} \\
    \midrule
    1 & The response is largely inaccurate, irrelevant, or incomplete, with poor language quality. It does not effectively address the question or may be incoherent. \\
    \midrule
    2 & The response contains significant inaccuracies or is missing key details. It may be unclear or poorly structured, and the language quality could be improved. \\
    \midrule
    3 & The response is generally accurate but may contain noticeable errors or omissions. It addresses the question with moderate clarity and completeness, but could be better structured or more detailed. \\
    \midrule
    4 & The response is mostly accurate and relevant, with a few minor errors or omissions. It is clear and well-structured but could benefit from slight improvements in detail or language quality. \\
    \midrule
    5 & The response is accurate, relevant, coherent, and complete, with excellent language quality. It answers the question thoroughly, clearly, and correctly, with no significant errors. \\
    \bottomrule
\end{tabular}
\caption{General scoring criteria for assessing the overall quality of responses by human judges. For each score, we provide guidelines to promote consistency and reliability in human evaluations.}
\label{tab:human_rules}
\end{table*}